\documentclass[letterpaper, 10 pt, journal, twoside]{IEEEtran}
\IEEEoverridecommandlockouts                              % This command is only needed if 
\usepackage[dvipdfmx]{graphicx}
\usepackage{epsfig} % for postscript graphics files
\usepackage{multirow}
\usepackage{color}
\usepackage{comment}
\usepackage{afterpage}
\usepackage{setspace}
\usepackage{amsmath}
\usepackage{algorithm}
\usepackage{overpic}
\usepackage[noend]{algorithmic}
\title{
Compensation for undefined behaviors during robot task execution by switching controllers depending on embedded dynamics in RNN
}

\author{Kanata Suzuki$^{1,2}$, Hiroki Mori$^{3}$, and Tetsuya Ogata$^{2,4}$
\thanks{Manuscript received: October, 14, 2020; Revised January, 12, 2021; Accepted February, 13, 2021.
This paper was recommended for publication by Editor Tamim Asfour upon evaluation of the Associate Editor and Reviewers’ comments.
This work was supported by JST, ACT-X Grant Number JPMJAX190I, Japan.} %Use only for final RAL version
\thanks{$^{1,2}$Kanata Suzuki is with Artificial Intelligence Laboratories, Fujitsu Laboratories LTD., Kanagawa 211-8588, Japan,
and also with Department of Intermedia Art and Science, School of Fundamental Science and Engineering, Waseda University, Tokyo 169-8555, Japan,
    {\tt\footnotesize suzuki.kanata@fujitsu.com}.}
\thanks{$^{3}$Hiroki Mori is with Institute for AI and Robotics, Future Robotics Organization, Waseda University, Tokyo 169-8555, Japan,
    {\tt\footnotesize mori@idr.ias.sci.waseda.ac.jp}.}
\thanks{$^{2,4}$Tetsuya Ogata is with Department of Intermedia Art and Science, School of Fundamental Science and Engineering, Waseda University,
and also with National Institute of Advanced Industrial Science and Technology, Tokyo 100-8921, Japan,
    {\tt\footnotesize ogata@waseda.jp}.}
\thanks{Digital Object Identifier (DOI): see top of this page.}
}

\begin{document}

\markboth{IEEE Robotics and Automation Letters. Preprint Version. Accepted February, 2021}
{Suzuki \MakeLowercase{\textit{et al.}}: Compensation for undefined behaviors by realizing embedded dynamics in RNN}

\maketitle
%\thispagestyle{empty}
%\pagestyle{empty}

%--------------------------------------------------------------------------
\begin{abstract}

Robotic applications require both correct task performance and compensation for undefined behaviors. Although deep learning is a promising approach to perform complex tasks, the response to undefined behaviors that are not reflected in the training dataset remains challenging. In a human--robot collaborative task, the robot may adopt an unexpected posture due to collisions and other unexpected events. Therefore, robots should be able to recover from disturbances for completing the execution of the intended task. We propose a compensation method for undefined behaviors by switching between two controllers. Specifically, the proposed method switches between learning-based and model-based controllers depending on the internal representation of a recurrent neural network that learns task dynamics. We applied the proposed method to a pick-and-place task and evaluated the compensation for undefined behaviors. Experimental results from simulations and on a real robot demonstrate the effectiveness and high performance of the proposed method.

\end{abstract}


\section{INTRODUCTION}

\IEEEPARstart{R}{obots} working with humans should correctly respond to situations that cannot be foreseen by developers. Although deep learning (DL) methods are available to generate motion trajectories (time-series data of joint angle positions) from end-to-end sensor information~\cite{E2E1}\cite{E2E2}, it is challenging to predict training data extrapolation. Even for self-supervised learning~\cite{E2E1}\cite{DRL3}\cite{DRL4} or generative adversarial imitation learning~\cite{GAN1}\cite{GAN2}, the generalization ability depends on the design of rewards or robot task motions. Thus, there is always a limit to the generalization ability of the DL method. In this study, the robot motion existing beyond the generalization ability range is called undefined behavior. Although many cases of undefined behavior exist, such as further generalization to novel tasks or environments, in this work, we studied managing undefined behaviors that interfere with the task execution, such as disturbances. To compensate for the decrease in the main controller's task performance, we propose an error recovery method to switch to another controller that controls undefined behaviors.

A typical example requiring compensation for undefined behaviors is an error recovery during human--robot collaborative works. Robots may express undefined behaviors due to collisions and other events. For a robot to neither damage the working environment nor represent a risk to humans in the collaboration environment, its controller should provide an appropriate posture upon the occurrence of undefined behaviors and restart the main task.

%-------------------- Fig --------------------
\setlength\textfloatsep{5pt}
\begin{figure}[t]
    \centering
    \includegraphics[width=8.2cm]{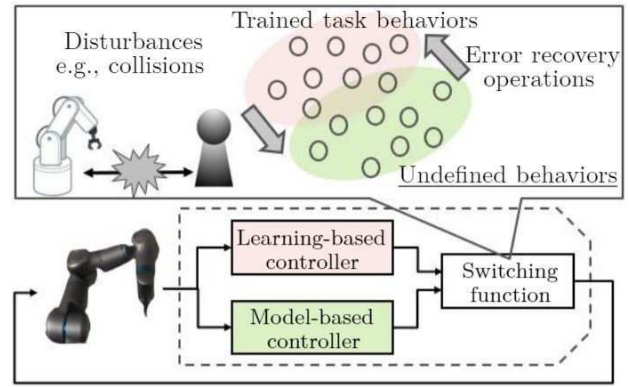}
    \caption{
Overview of our method to detect and compensate for undefined behaviors of a learning-based controller during task execution.
    }
\end{figure}

To switch between multiple controllers when the above error recovery operation, the undefined behaviors should be identified during task execution. This task can be considered anomaly detection. In anomaly detection methods for time-series data, outlier, change-point, and anomaly part detections exist. The anomaly part detection method is appropriate for the robot tasks dealt with in this study from the viewpoint of online detection. However, normal anomaly detection methods~\cite{Ano-detec2}\cite{Ano-detec3} rely on a detailed model that is independent of the controller. The task performance might decrease under the diverging prediction ability of the principal learning-based controller and anomaly detector. For controllers with different purposes to work together, generalization and compensation abilities (detecting and controlling for undefined behaviors) must be balanced. To prevent the above problem, our method detects undefined behaviors by evaluating the generalization ability of the main controller itself.

We propose a method to compensate for unexpected outputs of a learning-based controller by selecting a model-based controller when a robot expresses undefined behaviors. Note that we use the term ``model-based'' as a hand-crafted model, such as in modern control theory. To switch between the two controllers, the undefined behaviors should be identified during task execution. In this paper, we leverage the robot's sensorimotor experiences learned by a neuro-dynamical model. The proposed method determines whether the current trajectory belongs to the trained motion trajectories based on comparisons among previous trajectories in a recurrent neural network (RNN). By adding neurons to determine previous motion trajectories from the internal representation of the task-trained RNN, the learning-based method can detect undefined behaviors while maintaining the task performance. In addition, we adopt motion switching to design subtasks as attractors embedded in the RNN~\cite{Dy-Sw1}\cite{Dy-Sw2}. By combining the switching strategies, we incorporate the error recovery functions into the RNN without designing a separate anomaly detector. Our contributions are as follows:
\begin{itemize}
    \item We propose a method that compensates for undefined behaviors in a learning-based controller by using the internal representation of a task-trained RNN.
    \item We evaluate error recovery and the generalization ability of robot tasks under disturbances through simulations and experiments with a real robot.
\end{itemize}

\section{RELATED WORKS}

\subsection{Control Compensation for Undefined Behaviors}

Beyond robotics, various studies have addressed the combination of a high-efficiency and low-reliability model with a low-efficiency and safe model. In software engineering, Ortega et al.~\cite{Software-1} and Simard et al.~\cite{Software-2} analyzed a mixed system design of conventional deductive and inductive systems based on machine learning. The mixed system monitors the input and output from a machine learning method and incorporates exception processing, recovery processing, and multiplexing. Likewise, we apply this concept to robot control in the proposed method. Our method evaluates the trained task behaviors of the learning-based controller and implements error recovery through a model-based controller.

In reinforcement learning (RL) applied to robotics, a model-based controller is often adopted to support training. However, existing studies have neglected compensation for undefined behaviors~\cite{RL-Con-1}\cite{RL-Con-2}, and tasks are limited to motions that a model-based controller can be designed~\cite{RL-Con-3}. On the other hand, automatic driving with correction of the predicted trajectory to a target trajectory has been proposed using a learning-based controller~\cite{Onishi}. Although that method uses a model-based controller for augmenting training data, its performance in a real environment has not been evaluated. In control engineering, methods that combine machine learning and conventional control techniques have been developed. Sasaki et al.~\cite{r28-1} and Vamvoudakis et al.~\cite{r28-2} proposed a method to learn the feedback coefficient matrix of a linear quadratic regulator (LQR) through RL. Nevertheless, scarce research is available on the compensation for undefined behaviors of learning-based methods during real robot operation.

%------------------------------------- 
\subsection{Error Recovery during Robot Task Execution}

Error recovery during the performance of robot tasks has been extensively addressed in manufacturing. Previous studies~\cite{ER-d1}\cite{ER-d2} have been aimed to classify errors in layers, but this design results costly in practice. Although recovery processes have been derived using Petri nets~\cite{ER-l2}, switching strategies considering high-dimensional sensor information have not been devised. In control engineering, the control barrier function has been used for control that satisfies various requirements~\cite{B-con-1}\cite{B-con-2}. This function allows expressing collision avoidance constraints, but it may also constrain the task performance. A switching strategy proposed in \cite{Dy-Sw1} is based on the embedded dynamics of RNN and can provide an affordable design of error recovery. However, as recovery motion depends on training data, stable performance may be unfeasible.

We extend the methods in \cite{Dy-Sw1} and \cite{Dy-Sw2} by implementing controller switching during task execution. The proposed method compensates for undefined behaviors of an RNN by switching to a simple optimal controller. The model-based control guarantees optimality and convergence and does not rely on a training dataset.

%-------------------- Fig --------------------
\setlength\textfloatsep{5pt}
\begin{figure*}[t]
    \centering
    \includegraphics[width=16.4cm]{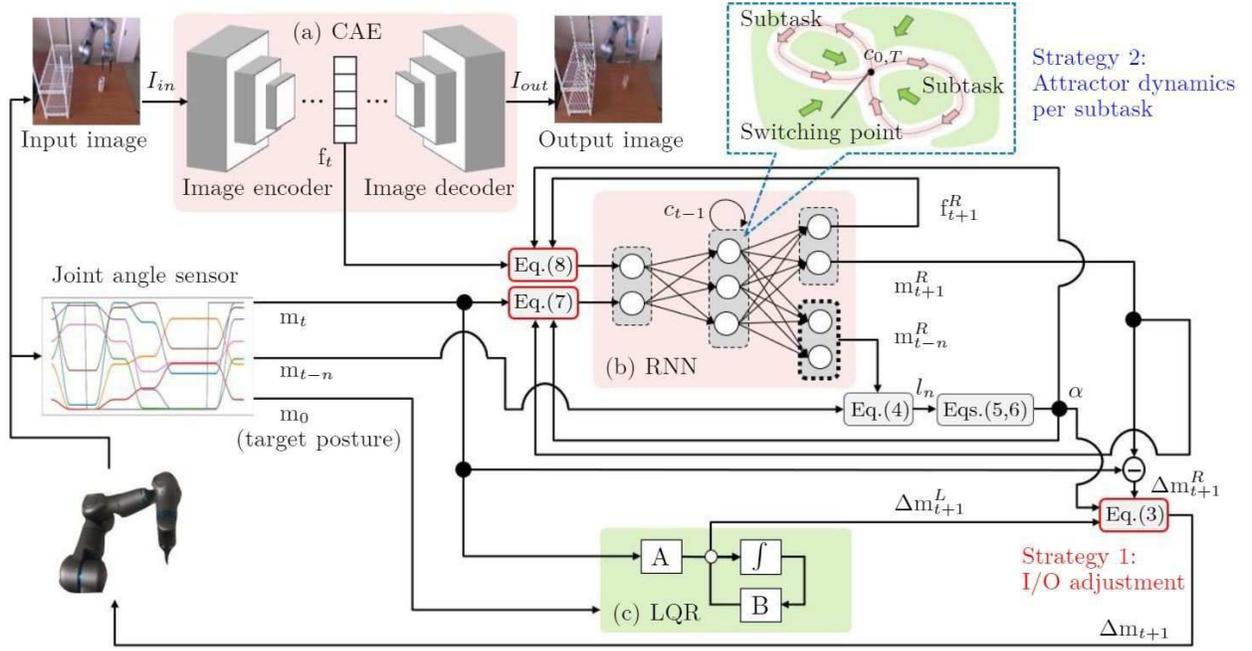}
    \caption{
Overview of the proposed method consisting of CAE for extracting (a) image features, (b) the RNN for learning the relationship between robot motion and image features, and (c) the LQR for controlling undefined behaviors. The method compares $\mbox{\boldmath m}_{t-n}$ with $\mbox{\boldmath m}_{t-n}^R$, and adjusts the RNN input/output (Strategy 1).The common internal RNN state in each subtask allows the RNN to execute an appropriate subtask after error recovery (Strategy 2).
}
\end{figure*}

\section{METHOD}

%The proposed method consists of image feature extractor, RNN, and one optimal controller (Fig. 2).
Our method consists of the learning-based controller (RNN and convolutional autoencoder (CAE)) and one optimal controller (Fig. 2).
%\textcolor{red}{Fig. 2 shows the details of each component and variables.}
In this section, we describe the learning procedure of our method and two switching strategies that leverage the internal RNN dynamics. 
%As shown in Fig. 2, the proposed method consists of two deep neural networks (image feature extractor and RNN) and one optimal controller. 

%------------------------------------- 
\subsection{Learning-based controller}

%The learning-based controller consists of an RNN and a convolutional autoencoder (CAE).
We train the models with sequences of joint angle and images as the robot's sensorimotor experiences. All sequences were acquired using a direct teaching.

%------------------------------------- 
\textbf{CAE: }In order to learn sensorimotor information properly, appropriate feature extraction from a high-dimensional image is important. First, we train the CAE, which is a sandglass-type multilayered neural network~\cite{AE} (Fig. 2(a)).
The model comprises fully connected, convolutional, and deconvolutional layers.
By training the CAE to provide output values that are equal to the input values, the model can extract low-dimensional image feature vectors in the middle layer.
When the CAE could reconstruct the image well, the encoded vectors reflected the relationship between the robot manipulator and object positions. The loss function is the mean squared error (MSE) between the output and input images, namely,
\begin{eqnarray}
    L_{CAE} = \frac{1}{N'} \sum_n^{N'} {(I_{in, n} - I_{out, n})^2},
\end{eqnarray}
where, $I_{in}$ and $I_{out}$ are input and output images, and $N'$ is a number of total images in the training dataset.

%------------------------------------- 
\textbf{RNN: }After training the CAE, we trained the RNN to learn the relationship between the robot motion and image features. By learning this relationship, RNN can predict the appropriate motion based on the current environment situation online. In the forward calculation, the RNN outputs the next state of robot joint angles $\mbox{\boldmath m}_{t+1}^R$ and image features $\mbox{\boldmath f}_{t+1}^R$ from the current state, $\mbox{\boldmath m}_t$ and $\mbox{\boldmath f}_t$ (Fig. 2(b)). Our method does not depend on the type of RNN, and we tried LSTM~\cite{LSTM}, GRU~\cite{GRU}, MTRNN~\cite{MTRNN} as the RNN function in our experiments (Table I).
%In forward calculation, the RNN outputs the next state of joint angles and image features from the current state of robot joint angles and image features; 
%\begin{eqnarray}
%    m_{t+1}^R, f_{t+1}^R = RNN(m_t, f_t, c_{t-1}), 
%\end{eqnarray}
%where, $m$ is robot joint angles, $f$ is image features, $c$ is the hidden state, $\bullet^R$ is the output value of RNN, $t$ is current time step, and $\bullet_t$ is the value at the time step $t$.

As our study aims to achieve error recovery in robot tasks, the RNN must restart the execution of a task from the middle or beginning after the robot recovers from an undefined behavior. To realize that, we use the data division method and the loss function proposed in \cite{Dy-Sw1}\cite{Dy-Sw2}.
We trained the RNN with subtask motion sequences that were obtained by dividing the original task sequences. 
The final and initial states of consecutive subtasks share the same posture, which represents the switching point of the corresponding subtask. 
For each subtask, the robot posture at the switching point is the initial posture of the task (Fig. 3). 
The loss function of the RNN is minimized for every training sequence, namely,
\begin{eqnarray}
L_{RNN}=\sum_{t=0}^{T-1} \left[ (\mbox{\boldmath m}_{t+1}^R-\hat{\mbox{\boldmath m}}_{t+1})^2 
%L_{RNN}=\sum_{t=0}^{T-1} \left[ (\mbox{\boldmath m}_{t+1}-\textcolor{red}{\hat{\mbox{\boldmath m}}_{t+1}})^2 
                        + (\mbox{\boldmath f}_{t+1}^R-\hat{\mbox{\boldmath f}}_{t+1})^2 \right] \nonumber \\
[-1.6mm]
                        + (\mbox{\boldmath c}_{0}-\mbox{\boldmath c}_{T})^2,
\end{eqnarray}
%, \mbox{\boldmath $\hat{m}$} and \mbox{\boldmath $\hat{f}$} are the target signals, and \mbox{\boldmath $c$} represents the internal RNN state. 
where $T$ is the sequence length, $\hat{\bullet}$ is the target value from training dataset, $\mbox{\boldmath m}$ are robot joint angles, $\mbox{\boldmath f}$ are the image features extracted by the trained CAE, $\mbox{\boldmath c}$ is the hidden state of the RNN, $\bullet^R$ is the output value of the RNN, $t$ is current timestep, and $\bullet_t$ is the value at the timestep $t$.
The first and second terms correspond to the MSEs between the outputs and target signals and between the initial and final RNN states in each subtask, respectively. 
By minimizing the loss function, the second term prevents bringing the initial and last hidden state values in each training sequence closer together. Since the current state ($\mbox{\boldmath m}_t$, $\mbox{\boldmath f}_t$) and the previous hidden state ($\mbox{\boldmath c}_{t-1}$) determine the RNN output, the RNN learns iterative sequences. Furthermore, because the initial hidden state value is common in all sequences, the RNN acquires subtasks as attractor dynamics with a common internal state~\cite{Dy-Sw1}\cite{Dy-Sw2}. This dynamical system allows the robot to switch between subtasks based on sensor information at a common internal state during motion generation.

%\vspace{-0.08mm}
%------------------------------------- 
\subsection{Training RNN with Model-based controller}

%We use an optimal model-based controller to control the robot when undefined behaviors occur. 
We use an optimal model-based controller to control undefined behaviors (Fig. 2(c)).
The controller recovers the robot posture to prevent task failure, operates at the joint level to set the control target, and uses the initial robot posture as the convergence target. 
In our experimental setting, we adopted an LQR, which is a simple yet widely used optimal controller~\cite{r28-1}\cite{r28-2}. 
The LQR controls joint angles linearly for the robot to directly return to an initial posture from the current posture. 
%Note that the LQR was used to illustrate the effectiveness of the proposed controller switching and more sophisticated model-based controllers can be adopted for difficult operations. 

As our method switches between controllers during task execution, the RNN should retain its generalization ability while using another controller. 
To this end, we embed the dynamics of the LQR into the RNN. 
The output difference of the optimal controller is connected to that of the RNN: %by using a skip connection~\cite{Res}:
\begin{eqnarray}
\Delta \mbox{\boldmath m}_{t+1} \leftarrow (1.0-\alpha) \Delta \mbox{\boldmath m}_{t+1}^R + \Delta \mbox{\boldmath m}_{t+1}^L,
\end{eqnarray}
where $\Delta \mbox{\boldmath m}_{t+1}$ is a command to the robot, $\bullet^R$ is the output value of RNN, $\bullet^L$ is the output value of LQR, and $\alpha$ is the coefficient for switching controllers during online operation.
In the training phase, we set $\alpha=0.0$ and the RNN learns the task motion while being inputted $\Delta \mbox{\boldmath m}_{t+1}^L$. The RNN becomes robust against perturbations caused by controller switching.
%The output difference of the optimal controller, $\Delta \mbox{\boldmath m}_{t+1}^L$, is connected to that of the RNN, $\Delta \mbox{\boldmath m}_{t+1}^R$, by using a skip connection~\cite{Res}: 
%By training the RNN for the total output difference of the two controllers, the model becomes robust against perturbations caused by controller switching. 

%------------------------------------- 
\subsection{Predicting Previous Motion Trajectories}

To perform switching, the proposed model must determine whether the robot is following a valid task trajectory. 
We trained additional neurons to determine a previous motion trajectory from the internal RNN representations. 
These neurons are connected to the middle context layer of the RNN and provide previous joint angles, $\mbox{\boldmath m}_{t-n}^R $, up to $N$ timesteps before the current timestep. 
Thus, the RNN predicts motion trajectories by reusing its internal transition obtained during training. 
From the previous motion trajectory, the error of the generated behavior can be calculated. 
The model minimizes MSE of the previous joint angles per training sequence: 
\begin{align}
L_{RNN,p} = \cfrac{1}{N}\sum_{n=1}^N l_n = \cfrac{1}{N}\sum_{n=1}^N (\mbox{\boldmath m}_{t-n}^R - \mbox{\boldmath m}_{t-n})^2.
\end{align}
In Fig. 2(b), the additional neurons are surrounded by a dotted black line.
%The RNN weight and bias for training the main tasks are fixed when training previous motion trajectories. 
The RNN weight and bias for training the main tasks are fixed when training previous motion trajectories. The RNN and AE have performed high-difficulty tasks such as flexible object manipulation~\cite{E2E2}\cite{Dy-Sw2}, and our method can add an error recovery function to those methods without decreasing their task performances.
%The RNN and AE have performed high-difficulty tasks such as flexible object manipulation~\cite{E2E2}\cite{Dy-Sw2}, and the proposed method can add an error recovery function to those methods without decreasing their task performances.

%------------------------------------- 
\subsection{Switching strategies}

%To switch between controllers while generating robot motions, the proposed method adopts two switching strategies: RNN input/output adjustment based on the determined previous motion trajectory (strategy 1) and designed dynamic systems in the RNN for the switching subtask (strategy 2). 
Our method adopts two switching strategies: RNN input/output adjustment based on the previous motion trajectory (strategy 1) and designed dynamical systems in the RNN for the switching subtask (strategy 2).

%------------------------------------- 
\textbf{Strategy 1: }
To switch between the controllers, the RNN prioritizes the model-based controller under undefined behaviors. 
The RNN determines whether the current posture corresponds to a trained motion based on the error of the previous and real motion trajectory. 
The error at timestep $n$, $l_n$, increases if the current robot state reflects an undefined behavior. 
The input/output of the RNN are adjusted according to coefficient $\alpha$ derived from $l_n$: 
\begin{eqnarray}
\alpha_n = \frac{1}{1+exp(- \beta (l_n - \gamma \overline{l_n}))}, \\ 
[1.5mm]
\alpha = \max \{ \alpha_n \, | \, 0 \leq n \leq N \}, \\
\mbox{\boldmath m}_t \leftarrow (1.0 - \alpha) \mbox{\boldmath m}_t + \alpha \mbox{\boldmath m}_t^R, \\
\mbox{\boldmath f}_t \leftarrow (1.0 - \alpha) \mbox{\boldmath f}_t + \alpha \mbox{\boldmath f}_t^R,
\end{eqnarray}
where $\overline{l_n}$ is the average value of $l_n$ across five trials of the trained task sequences, and $\alpha$ is the maximum value of $\alpha_n$.
$\beta$ and $\gamma$ are parameters to adjust the sensitivity of controller switching.
Eq. (5) is a Sigmoid function, $\beta$ represents its gradient, and $\gamma$ represents its bias.
We set $\beta=$30 and $\gamma=$3, depending on the simulation results on the validation dataset. When $\alpha$ increases, $\Delta m_{t+1}^R$ is restricted and the model switches to the model-based controller (Eq. (3)).
Furthermore, the method adjusts the RNN input in a closed-loop, depending on $\alpha$ to suppress the effects of disturbances (Eqs. (7,8)).

%------------------------------------- 
\textbf{Strategy 2: }
After the robot recovers the initial posture using the model-based controller, it should appropriately resume the task execution. 
We use the switching points from the RNN. 
As described in Section III-A, the initial posture of the robot is a switching point. 
When the robot returns to the initial posture of a trained behavior, $\alpha$ decreases, and the RNN output is prioritized. 
The RNN can switch to the subtask at the switching point that is in an equilibrium state, where the input for starting each subtask is obtained. 
This enables the robot to resume the task execution depending on the input camera information after error recovery.

\section{Experiments}

%-------------------- Fig --------------------
\begin{figure*}[t]
    \centering
    \includegraphics[width=16.4cm]{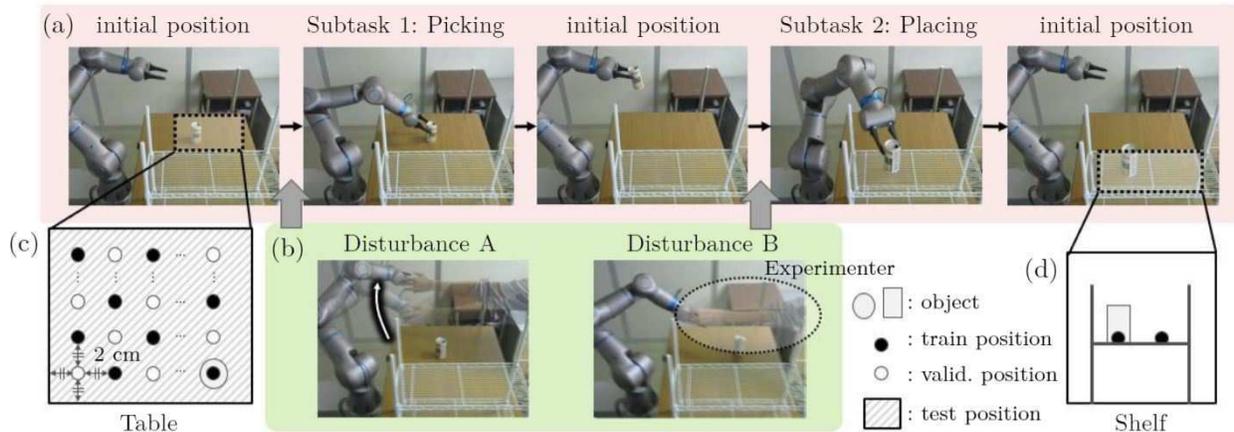}
    \caption{
Pick-and-place task comprising two subtasks.
The robot picks a can from a desk and places it in a vacant spot on a shelf.
During experiments, we induced two types of trajectory disturbances.
    }
\end{figure*}

%------------------------------------- 
\subsection{Task Design}

We designed the pick-and-place task shown in Fig. 3(a) for execution using the Torobo ARM~\cite{torobo}. 
The task comprises two subtasks: 1) the robot picks a can from a desk, and then 2) places it in a vacant spot on a shelf. 
After each subtask, the robot returns to its initial posture. 
The camera images acquired during task execution reflect the object position and whether the robot is grasping the object. 
The robot should execute the appropriate subtask depending on the camera information. 
We prepared 100 motion sequences for subtask 1 with different object positions for picking. 
Specifically, the object was placed at intervals of 2 cm on the table. 
For subtask 2, we prepared two motion sequences with different spots for placing. 
As a combination of subtasks, 100(picking)$\times$2(placing)$=$200 sequences were possible, and we used 50$\times$2$=$100 sequences for training (Fig. 3(c,d)).
The remaining sequences were used as the validation dataset.

We collected the training robot motion data through direct teaching. The training data for both images and joint angles were sampled at ten frames per second.
Subtasks 1 and 2 consisted of 161 steps and 192 steps on average, respectively.
We obtained seven-dimensional joint angles and one-dimensional gripper open/close information. 
The input images to the CAE were downsampled to 112$\times$112 pixels across the RGB channels. 
The images were acquired from a fixed camera pointing to the robot and the table.

%------------------------------------- 
\subsection{Training Setup}

Table I lists the parameters of the models in the proposed method. 
The CAE extracts 30-dimensional image features from the input images. 
To demonstrate that the proposed method does not depend on the type of RNN, we prepared the three RNNs listed in Table I: LSTM~\cite{LSTM}, GRU~\cite{GRU}, and multiple timescale RNN (MTRNN~\cite{MTRNN}). 
These RNNs have 38-dimensional input/output neurons to accept image features, joint angles, and gripper values. 
For error recovery, we adopted an LQR that provides optimal control by linearly regulating the joint angles to converge to the initial posture of the subtasks. 
We adjusted the LQR for its output to not exceed the RNN scaling range.

The RNN input was scaled to [$-$1.0, 1.0]. 
For training RNN and CAE, we used an Adam optimizer ($\alpha_{Ad}=0.001$, $\beta_{1,Ad}=0.9$, and $\beta_{2,Ad}=0.999$). 
We augmented the task sequences by adding Gaussian noise and applying color augmentation to increase the robustness of the models.

%-------------------- Table --------------------
\begin{table}[t]
    \centering
    \begin{tabular}{c|c}
        \multicolumn{2}{c}{TABLE I: Structures of the networks} \\
        \hline
        Network         & Dims \\
        \hline \hline
        CAE$^{1,2}$  & input@3chs - convs@(32-64-128)chs - \\ 
                   & full@1000 - full@30 - full@1000 - \\ 
                   & dconvs@(128-64-32)chs - output@3chs \\ 
        \hline
        LSTM$^{3}$   & $IO$@38 - $hidden$@250 (1 layer) \\ 
        \hline 
        GRU$^{3}$     & $IO$@38 - $hidden$@250 (1 layer) \\ 
        \hline 
        MTRNN$^{3}$ & $IO$@38 - $C_f$@80($\tau$:2) - $C_s$@20($\tau$:30) \\ 
        \hline 
    \end{tabular}
    \begin{flushleft}
        1) all conv \& dconv are kernel 6, stride 2, and padding 1. \\
        2) all layers have batch norm. and relu activation. \\
        3) all layers have tanh activation. \\
    \end{flushleft}
\end{table}

%------------------------------------- 
\subsection{Performance Evaluation}

We verified the compensation ability of the proposed model for undefined behaviors in a real robot. The robot performed a task under disturbance applied by an experimenter (Fig. 3(b)). We set two types of disturbances assuming collisions during task execution. When the robot approached the picking position or placing position for the object, we either moved the robot posture to an undefined posture that was not included in the training data (disturbance A) or stopped the robot motion for a period (disturbance B). The transition posture under disturbance A was determined randomly within a range that prevented damaging the experimental environment. The robot posture was maintained for approximately 10 timesteps after the disturbances. Thus, the method had to perform the task while handling disturbances.

We conducted simulator experiments to compare the proposed method with conventional time-series anomaly detection methods, $k$-nearest neighbors (kNN~\cite{kNN}) and subspace method (SM~\cite{SM}). We adopted kNN and SM because they are used in various anomaly part detection tasks and are easy to implement. Furthermore, since the kNN and SM are not required to build predictive models meticulously, they are suitable for evaluating the proposed method regarding the balance of compensation and generalization abilities. We created a partial time-series dataset from the training data by shifting a time window, which size was set to 10. By using the time-series dataset, we built kNN and SM models for each joint angle. In the kNN, we set $k=1$. In the SM, we used the two axes with the highest contribution. In the test phase, we set $N=$window size and evaluated the past time-series online. The kNN and SM output values are treated as $l$ in Eq.(6). We then adopted the one with the highest anomaly in the joint angles as $\alpha$. $\beta$ and $\gamma$ for kNN and SM were set to their best values after some trials on the validation dataset. We judged task success based on both the successful controller switching and generate subtask motion after error recovery operation.

In addition, we conducted real robot experiments. We considered the task to be successful if the robot picked the object and placed it in the vacant spot on the shelf through the combination of subtask operations. Furthermore, the generalization ability of the picking task was evaluated by placing the object positions randomly at unlearned positions 1.0--2.0 cm away from the trained positions.

\section{Results and Discussion}

%------------------------------------- 
\subsection{Estimation of Previous Motion Trajectories}

First, we verified the ability of the RNN to determine previous motion trajectories. Figure 4 shows the training loss of the joint angles according to the previous timestep. Although all RNNs successfully learned previous motion trajectories, a large estimation horizon ($N$) undermines the estimation performance. This indicates the immediately preceding dynamics needed to perform the task are embedded in the internal RNN state. Regarding the sensitivity to disturbances, it is not desirable to use unnecessarily distant previous predictions. Hence, we used up to 10 previous steps ($N=10$) to detect undefined behaviors. The prediction performance of the GRU was particularly poor among the RNNs, consequently providing a low detection performance of undefined behaviors.

%------------------------------------- 
\subsection{Simulator Experiments}

We conducted simulations to investigate the switching ability of the proposed method. Figure 5 shows examples of sequences generated by closed-looped LSTM. Figure 5(a) shows the sequence obtained without disturbances, whereas Figs. 5(b) and (c) show the sequences obtained under disturbances A and B, respectively. The results indicate that the proposed method restricted the output of the RNN after the disturbance by multiplying the $\alpha$ value. If controller switching did not perform, the robot might have moved unreasonably due to $\Delta \mbox{\boldmath m}^R$ altered by the disturbance. The proposed method suitably handles undefined behaviors during task execution.

%-------------------- Fig --------------------
\begin{figure}[t]
    \centering
    \includegraphics[width=8.4cm]{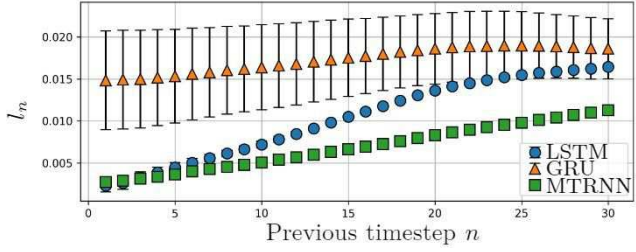}
    \caption{
Training loss of joint angles according to previous timestep. 
    }
\end{figure}

%-------------------- Table --------------------
\begin{table}[t]
    \centering
    \begin{tabular}{c|ccc}
        \multicolumn{4}{c}{TABLE II: Success rates for controller switching} \\
        \hline
               & No dist.       & Dist. A        & Dist. B \\
        \hline     \hline
        kNN    & 90.8$\pm{3.1}\%$ & 83.3$\pm{5.1}\%$ & 81.7$\pm{4.7}\%$ \\ 
        \hline
        SM     & 91.7$\pm{4.7}\%$   & 84.2$\pm{4.2}\%$ & 80.8$\pm{5.1}\%$ \\ 
        \hline
        Ours (LSTM)  & 100.0$\pm{0.0}\%$  & 95.8$\pm{2.4}\%$ & 94.2$\pm{3.1}\%$ \\  
        \hline
    \end{tabular}
\end{table}

%-------------------- Fig --------------------
\begin{figure}[t]
    \centering
    \includegraphics[width=8.6cm]{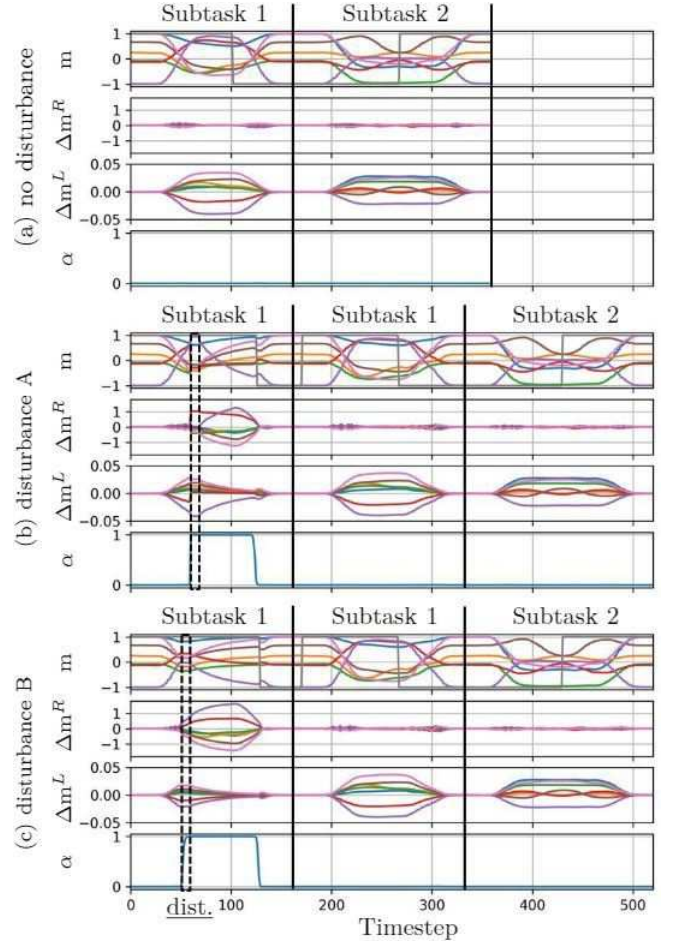}
    \caption{
Sequences generated by closed-looped LSTM under (a) no disturbance, (b) disturbance A, and (c) disturbance B. 
The lines of $\mbox{\boldmath m}$, $\Delta \mbox{\boldmath m}^R$, and $\Delta \mbox{\boldmath m}^L$ indicate joint angle values. 
    }
\end{figure}

Table II lists the success rates of our method, kNN, and SM. Each method generated 40 validation sequences three times using LSTM trained with random seeds. Successful controller switching was determined by the generated sequence becoming $\alpha=1$ under the disturbance and $\alpha=0$ at the initial robot posture after the LQR operation. The proposed method outperforms the conventional methods under all the conditions regarding the success rate. In particular, the conventional methods provide a low success rate even without disturbance. These methods did not provide generalization for the training data and judged correct motion as anomalies, leading to low task performance. In contrast, our method correctly switched the controllers under no disturbance condition.

As described in Section I, the detection performance of the external detector should be matched with the main controller. When diverse and large training datasets, proper parameter design of learning-based detector~\cite{Ano-detec2}\cite{Ano-detec3} is difficult. Alternatively, prediction can be used instead of previous trajectory estimation in the same structure of our method~\cite{Ano-detec2}. In this case, the RNN predicts some steps ahead and compares the predicted and actual trajectories. However, that approach fails to respond to sudden environmental changes online because it requires predicting the reference trajectory in advance. Overall, our method is effective for the online detection of undefined behaviors when using the learning-based controller.

%------------------------------------- 
\subsection{Real Robot Experiments}
%-------------------- Fig --------------------
\begin{figure*}[t]
    \centering
    \includegraphics[width=16.5cm]{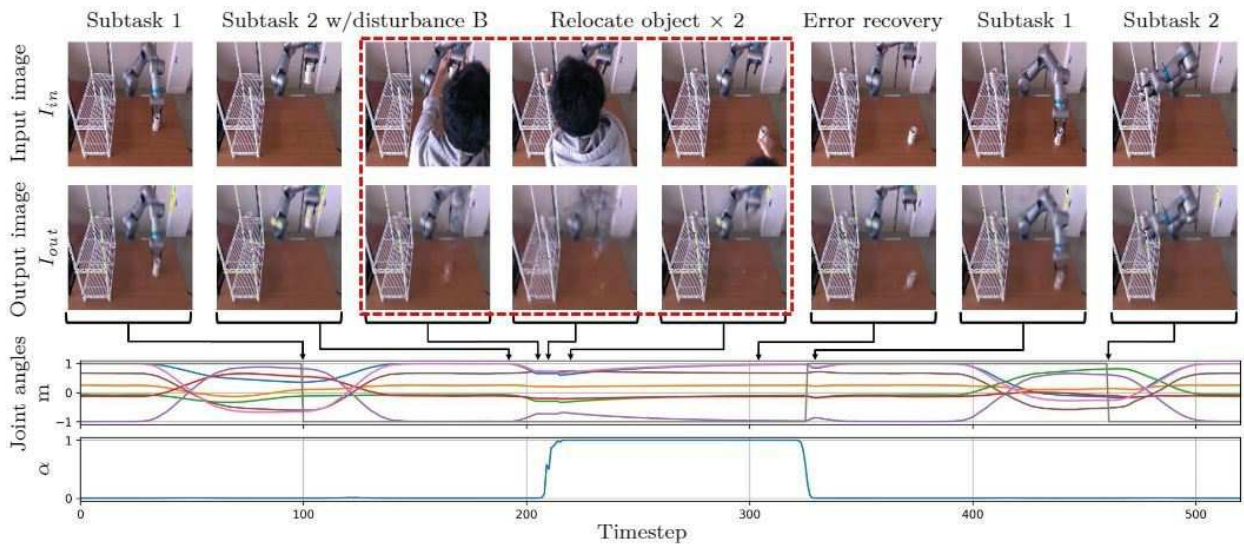}
    \caption{
Generated behaviors under disturbance B during execution of subtask 2. 
The output images were reconstructed from $\mbox{\boldmath f}$ by the CAE.
    }
\end{figure*}

To evaluate the error recovery and generalization, the robot was controlled to execute the pick-and-place task at untrained object positions. Figure 6 shows a generated motion trajectory under disturbance B. During the execution of subtask 2, the experimenter stopped the robot motion. Then, the can was removed from the robot gripper and relocated in different positions. The model switched the controller from the RNN to the LQR for the robot to return to its initial posture. Then, the robot restarted the task execution based on the camera information, and subtasks 1 and 2 proceeded depending on the new object positions. The proposed method enables the robot to perform both error recovery for undefined behaviors and task execution for untrained object positions. Although we used only two subtasks in our experiment, the method can manage multiple subtasks in reality (shown in \cite{Dy-Sw1}).

Sensitivity is an important parameter during controller switching. 
Unlike simulations, real-world perturbations present complex dynamics that are fed to the RNN during the experiments. 
As the proposed method determines the error over a time series, it can perform the task without accidentally switching to the model-based control even under small perturbations. 
In addition, although camera occlusion was caused when the disturbance (red frame in Fig. 6), the robot generated appropriate motion.

%-------------------- Table --------------------
\begin{table}[t]
    \centering
    \begin{tabular}{c|ccc}
        \multicolumn{4}{c}{TABLE III: Success rates in real robot experiments} \\ 
        \hline
              & No dist.       & Dist. A      & Dist. B \\ 
        \hline     \hline
        LSTM  & 74.2$\pm{4.7}\%$ & 71.2$\pm{6.6}\%$ & 70.0$\pm{5.4}\%$ \\ 
        \hline
        GRU   & 72.5$\pm{6.1}\%$ & 68.3$\pm{6.2}\%$ & 67.5$\pm{3.5}\%$ \\ 
        \hline
        MTRNN & 73.3$\pm{5.1}\%$ & 70.0$\pm{4.1}\%$ & 69.2$\pm{4.2}\%$ \\ 
        \hline
    \end{tabular}
\end{table}

Table III lists the task success rate of each type of RNN in the experiments. The disturbance was applied at random instants during the execution of subtask 1 or 2, and 40 trials were conducted for each RNN trained with three random seeds. Compared with no disturbance, most error-recovery operations were successfully performed under disturbances A and B, but the success rate was lower under disturbance B than under disturbance A. This is because our method was not able to properly identify the undefined behavior when the task execution was stopped, as the disturbance did not considerably alter the original trajectory. However, most task failures were because of subtask 1 rather than controller switching. Since our training data were smaller than normal DL methods, the model's prediction accuracies could be insufficient. Such problems can be solved by collecting more training data.

Comparing the three types of RNNs, while the success rate of LSTM was the highest under all the disturbance conditions, that of GRU was slightly lower under any disturbance. This may be attributable to the training results for determining previous motion trajectories (Table II). However, the accuracy was similar among the RNNs, suggesting that the performance of our method does not depend on the adopted learning-based controller.

Our method identified undefined behaviors from the RNN's internal representation. Therefore, the proposed method evaluates the RNN's generalization ability and uses the RNN control function within the appropriate generalization region (interpolation of the dataset), and the LQR control function otherwise (extrapolation of the dataset). Furthermore, the RNN can predict extrapolated but newly generalized situations (test object positions outside the training positions in Fig. 3(c)). We have shown that the error recovery function allows the method to return the robot's state to interpolating training data. Our research was the first step in extending the previous methods~\cite{E2E2}\cite{Dy-Sw2} to control undefined behaviors, and the experiments were only for limited training data extrapolation. For more advanced motion generation for undefined behaviors, we plan to experiment in future work.

%-------------------- Fig --------------------
\begin{figure}[t]
    \centering
    \includegraphics[width=7.7cm]{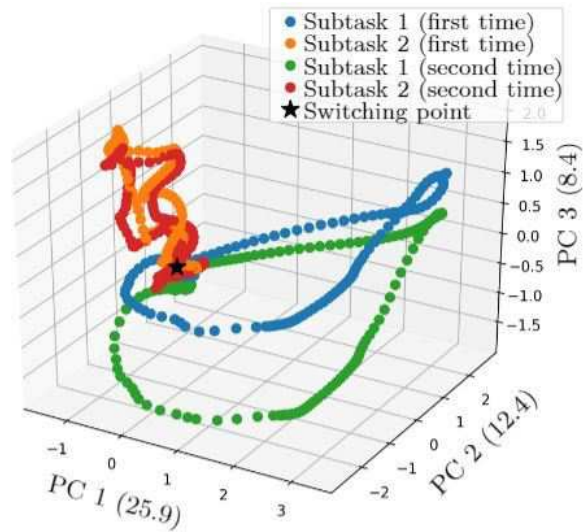}
    \caption{
Internal LSTM states visualized by PCA. The contribution ratios are PC1: 25.9\%, PC2: 12.4\%, and PC3: 8.4\%.
    }
\end{figure}

\subsection{Analysis of Internal RNN States}

At last, we visualized the internal LSTM states by using principal component analysis (PCA). The visualized dynamics indicate the generated motion shown in Fig. 6. The internal LSTM states form an attractor structure for each subtask, and the attractors share a common internal state, the switching point. After a disturbance, the LSTM continues to predict the task motion as an attractor that repeats the subtask dynamics by changing its input to a closed-loop structure. When the robot returns to the initial posture by LQR and the internal LSTM state returns to the switching point, the model transitions to the appropriate subtask depending on the image information at the switching point. The subtle differences in transition between the first and second subtasks correspond to the object positions. Therefore, our method can select subtasks based on the designed dynamical system while maintaining the switching ability of the controllers.

\section{Conclusion}

We proposed a method to compensate for undefined behaviors in a learning-based controller by switching to a model-based controller. Our method predicts previous motion trajectories from embedded dynamics in the RNN and determines whether the current trajectory corresponds to a trained motion. By reusing the embedded dynamics in the RNN, the method identifies undefined behaviors while maintaining the task performance. We demonstrated that the robot can successfully complete pick-and-place tasks even after motion interruption and disturbances.

%--------------------------------------------------------------------------

\end{document}